\documentclass[conference]{IEEEtran}
\IEEEoverridecommandlockouts

\usepackage{cite}
\usepackage{amsmath,amssymb,amsfonts}
\usepackage{algorithmic}
\usepackage{graphicx}
\usepackage{textcomp}
\usepackage{xcolor}
\usepackage{booktabs}
\usepackage{multirow}
\def\BibTeX{{\rm B\kern-.05em{\sc i\kern-.025em b}\kern-.08em
    T\kern-.1667em\lower.7ex\hbox{E}\kern-.125emX}}
\begin{document}

\title{An Attentive Dual-Encoder Framework
Leveraging Multimodal Visual and Semantic
Information for Automatic OSAHS Diagnosis\\
\thanks{
Published as a conference paper at ICASSP 2025
}
\thanks{
\IEEEauthorrefmark{1}This is to indicate the equal contribution.
}
\thanks{\IEEEauthorrefmark{2}This is to indicate the corresponding author. Email: qiuxihe@sues.edu.cn}
\thanks{This work is supported by Shanghai Committee of Science and Technology ``Science and Technology Innovation Action Plan'' Natural Science Foundation of Shanghai (23ZR1425400), the National Natural Science Foundation of China (62102241), Eye \& ENT Hospital's double priority project A (YGJC026 to Dr Wei and Dr Huang).}
}
\author{
    \IEEEauthorblockN{
        \textbf{Yingchen Wei}\textsuperscript{1}$^,$\IEEEauthorrefmark{1},
       \textbf{Xihe Qiu}\textsuperscript{1}$^,$\IEEEauthorrefmark{1}$^,$\IEEEauthorrefmark{2},
        \textbf{Xiaoyu Tan}\textsuperscript{2}$^,$\IEEEauthorrefmark{1},
        \textbf{Jingjing Huang}\textsuperscript{3},
        \textbf{Wei Chu}\textsuperscript{2},\\
        \textbf{Yinghui Xu}\textsuperscript{2},
        \textbf{Yuan Qi}\textsuperscript{4}}
 
    \IEEEauthorblockA{\textsuperscript{1}School of Electronic and Electrical Engineering, Shanghai University of Engineering Science, Shanghai, China\\
                      \textsuperscript{2}INF Technology (shanghai) Co., Ltd., Shanghai, China\\
                      \textsuperscript{3}ENT institute and Department of Otorhinolaryngology, Eye \& ENT Hospital, Fudan University, Shanghai, China\\
                      \textsuperscript{4}Artificial Intelligence Innovation and Incubation Institute, Fudan University, Shanghai, China}
}

\maketitle

\begin{abstract}
Obstructive sleep apnea-hypopnea syndrome (OSAHS) is a common sleep disorder caused by upper airway blockage, leading to oxygen deprivation and disrupted sleep. Traditional diagnosis using polysomnography (PSG) is expensive, time-consuming, and uncomfortable. Existing deep learning methods using facial image analysis lack accuracy due to poor facial feature capture and limited sample sizes. To address this, we propose a multimodal dual encoder model that integrates visual and language inputs for automated OSAHS diagnosis. The model balances data using randomOverSampler (ROS), extracts key facial features with attention grids, and converts basic physiological data into meaningful text. Cross attention combines image and text data for better feature extraction, and ordered regression loss ensures stable learning. Our approach improves diagnostic efficiency and accuracy, achieving 91.3\% top-1 accuracy in a four class severity classification task, demonstrating state of the art performance. Code is available at https://github.com/luboyan6/VTA-OSAHS.
\end{abstract}


\section{Introduction}
\begin{figure}[h]  
  \centering
\includegraphics[width=255pt]{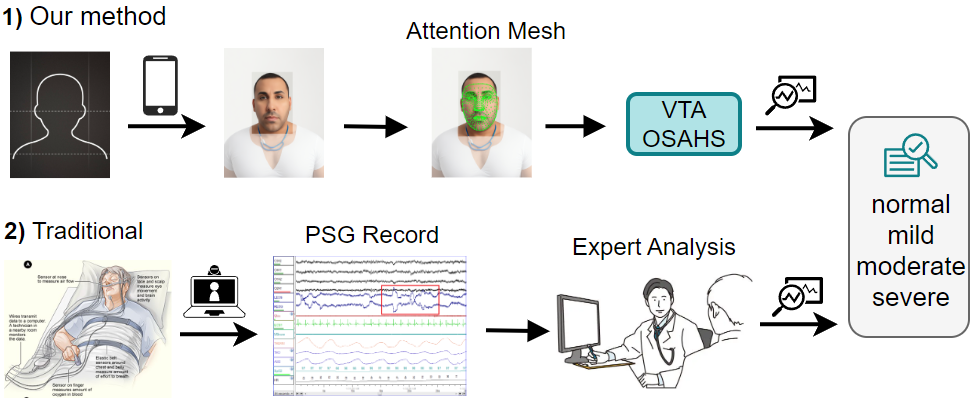} 
  \caption{A potential clinical alternative has been proposed to reduce the use of PSG for diagnosing OSAHS. This method conserves clinical resources, reduces patient waiting times, and cuts down on diagnostic costs.}
  \label{motive-fig}
\end{figure}

Obstructive sleep apnea-hypopnea syndrome (OSAHS) \cite{bib1} affects about 27\% of adults \cite{bib2}, causing poor sleep, daytime dysfunction, and higher risks of cardiovascular diseases and diabetes \cite{bib3}. The standard diagnostic method, polysomnography (PSG) \cite{bib4}, is complex, costly, and uncomfortable, requiring multi-channel monitoring (EEG, ECG, heart rate \cite{bib5}) and trained technicians (Fig. \ref{motive-fig}). Data-driven methods for automated OSAHS diagnosis can improve efficiency and reduce costs. Facial features like a flat nasal bridge, wide jawbone, thick neck, and mandibular retrognathia correlate with OSAHS severity \cite{bib7}, providing visual indicators of airway obstruction and sleep disturbances. Deep learning can analyze these features for early diagnosis and personalized treatment. While methods like SVM, logistic regression, and linear discriminant analysis have been used for this purpose \cite{bib8, bib9, bib10}, deep learning faces challenges with feature capture and single modal data limitations.

To address these issues, we propose VTA-OSAHS, a multimodal dual-encoder framework that integrates visual and semantic data for OSAHS diagnosis. It combines facial features and basic physiological data through an image encoder, text encoder, and multi-modal fusion module. The image encoder uses the attention mesh method \cite{bib11} to extract key facial points and stochastic gates \cite{bib12} for feature selection. The text encoder encodes physiological data (e.g., gender, height, weight) using Clinical BERT \cite{bib13}. A cross-attention mechanism and residual connections integrate visual and text data, enhancing model performance \cite{bib14}. Evaluation on a clinical dataset showed that VTA-OSAHS achieved 91.3\% accuracy in diagnosing OSAHS severity across four categories (normal, mild, moderate, severe) and a high AUC of 95.6\%, outperforming state-of-the-art methods.

Our key contributions are as follows: (1) Introducing VTA-OSAHS, a multimodal framework for diagnosing OSAHS severity by combining visual and language data, and using a pre-trained language model to extract key information from basic physiological data for improved classification accuracy; (2) Developing a visual encoder that focuses on specific facial features associated with OSAHS, employing attention mesh and stochastic gates for better clinical decision alignment; (3) Implementing a data pre-processing strategy to handle imbalanced samples and ordinal classification, using randomOverSampler (ROS) \cite{bib17} and an ordinal regression loss function \cite{bib18} to enhance accuracy and robustness; (4) Demonstrating state of the art performance on a real world clinical dataset, achieving human-level diagnostic accuracy and outperforming existing methods.

\section{Method}
\begin{figure}[h]
  \centering
  \includegraphics[width=255pt]{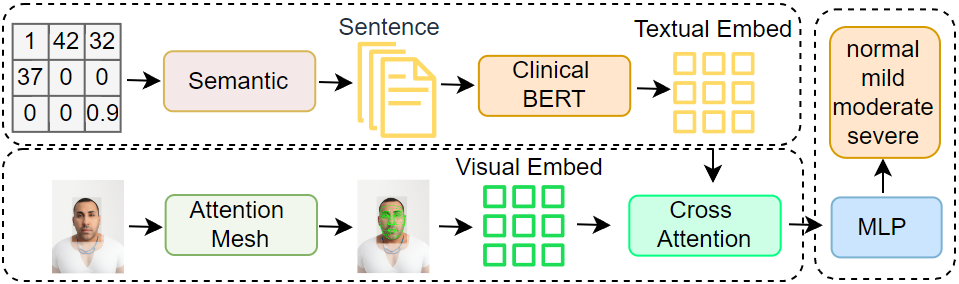}  
  \caption{The overall framework of our proposed method. It consists of an image encoder, a text encoder, and a multi-modal fusion module. Sentence: This 37-year-old male has a neck circumference of 42cm, a waist to hip ratio of 0.9, a body mass index of 32, indicating that he is obesity, and not history of hypertension, diabetes, heart disease, and hyperlipidemia.}
  \label{fig:model-fig}
\end{figure}

The proposed model integrates three components: an image encoder, a text encoder, and a multi-modal fusion module. It processes a patient's image and preprocessed patients' basic physiological data, encoding them into image, text, and combined image-text vectors. Using cross attention mechanisms, it captures contextual relationships between these modalities. The image encoder uses the Attention Mesh (AM) facial keypoint selector, and the text encoder is based on the Clinical BERT model. The fusion module combines image and text features, enhancing multi-modal understanding and improving task performance. The structure of the overall model for our proposed method is depicted in Fig. \ref{fig:model-fig}.

\subsection{Input of Image and Text Encoders}
The AM technique converts a patient's facial keypoints into a three dimensional vector. The model processes 256x256 pixel images to generate a 64x64 feature map, which is divided into submodules. One submodule predicts the 3D coordinates of 468 facial landmarks, critical for identifying key features in OSAHS patients, while the other predicts region landmarks using attention mechanisms. These vectors are then projected and flattened into a one-dimensional vector for block embeddings. The model, represented by AM, maps the input image to a facial mesh $\theta$ via an affine transformation, $\theta = AM(I)$, enabling geometric transformations such as scaling, rotation, and translation.
\begin{equation}
    \begin{array}{c}
    \theta
    \end{array}
    =
    \left[
    \begin{array}{ccc}
    x_{x} & sh_{x} & t_{x} \\
    sh_{y} & s_{y} & t_{y}  
    \end{array}
    \right]
\end{equation}
The affine transformation can be determined either through supervised prediction of matrix parameters or by deriving them from the facial mesh submodule's output. The feature extraction process, represented as $F(\theta)$, focuses on identifying key points relevant to OSAHS from the facial mesh.

After extracting facial key points using the AM technique, stochastic gates are employed for feature selection to enhance learning efficiency and reduce redundancy. These gates introduce a binary random variable, $G$, determining the activation of neurons. The network adjusts $G$ during training to minimize prediction errors. To ensure differentiability, the Gumbel-softmax \cite{bib34} technique is used to transform $G$ into a continuous probability distribution.

The output $Y_i^{(l)}$ of the $i$-th neuron in the $l$-th layer of a neural network is defined based on the input $X^{(l)}\in \mathbb{R}^{n_l \times m}$, where $n_l$ is the number of neurons in that layer and $m$ is the number of input features.
\begin{equation}
X_{i}^{(l)}= \sigma(\sum_{j=1}^{m}G_{i,j}^{(l)}\cdot w_{i,j}^{(l)}\cdot X_{j}^{(l)} )
\end{equation}
The weight $w_{i,j}^{(l)}$ connects the $j$-th input feature to the $i$-th neuron, and $\sigma(\cdot)$ is the activation function. The random gate variable $G_{i,j}^{(l)}$ indicates whether the $i$-th neuron in the $l$-th layer uses the $j$-th input feature. To make $G$ differentiable, Gumbel-softmax converts it into a probability distribution, $P(G)$, where $P_{i,j}^{(l)}$ denotes the probability of the $i$-th neuron using the $j$-th feature.

\begin{equation}
\mathrm{P}_{\mathrm{i}, \mathrm{j}}^{(l)} = \frac{\exp \left(\left(\log \left(\mathrm{G}_{\mathrm{i}, \mathrm{j}}^{(l)}\right)+\mathrm{g}_{\mathrm{i}, \mathrm{j}}^{(\mathrm{l})}\right) / \tau\right)}{\sum_{\mathrm{k}=1}^{\mathrm{m}} \exp \left(\left(\log \left(\mathrm{G}_{\mathrm{i}, \mathrm{k}}^{(l)}\right)+\mathrm{g}_{\mathrm{i}, \mathrm{k}}^{(l)}\right) / \tau\right)}
\end{equation}
The hyperparameter $\tau$ controls temperature, while Gumbel noise $g_{i,j}^{(l)}$ introduces randomness, enabling flexible feature selection based on the probability distribution $P(G)$. This reduces reliance on specific features and minimizes redundancy. Dimensionality reduction is applied to the feature sequence to select the optimal subset, preserving classification performance. A learnable token, $V_{cls}$, is added, resulting in the image input representation: $I = \{V_{cls},V_{1},...,V_{n}\}$.

We focused on key physiological factors related to the diagnosis of OSAHS, including gender, age, neck circumference, BMI, waist to hip ratio (WHR), and comorbidities like hypertension, diabetes, heart disease, and hyperlipidemia, which are readily available from outpatient data. However, raw numerical values lack semantic context. To address this, we converted these features into meaningful sentences and encoded them using Clinical BERT, adding a learnable special token $W_{cls}$, resulting in a text-based input representation: $T = \{W_{cls},W_{1},...,W_{n}\}$.

\subsection{Multi-Modal Fusion}
In multi-modal tasks, understanding modality interactions is essential. To integrate image and physiological data, encoded using attention mesh and Clinical BERT, a cross-attention mechanism is employed. This mechanism uses two sets of feature vectors, one for text input ($T$) and one for image input ($I$), to compute their similarity and determine feature correlations for effective fusion.

For model weight allocation, learnable weight matrices $W_{q}$, $W_{k}$, and $W_{v}$ project input features into query, key, and value spaces: $Q = IW_{q}$, $K = TW_{k}$, and $V = TW_{v}$. These matrices are learned during training to capture relevant features for cross-modal interaction. The mechanism then computes the scaled dot-product attention between query and key vectors to derive attention scores:
\begin{equation}
\operatorname{Attention}(Q, K, V)=\operatorname{softmax}\left(\frac{Q K^{\top}}{\sqrt{d_{k}}}\right) V
\end{equation}

The softmax function converts these scores into attention weights, creating a probability distribution over text features for each image. This cross-attention mechanism enhances the integration of OSAHS image and text data, improving model performance.

\subsection{Output Layer}
We applied the ROS algorithm to balance the dataset. The original dataset $D=\left \{(x_1, y_1),(x_2, y_2),\ldots,(x_N, y_N)\right \}$ contains samples with features $x_i$ and labels $y_i$. We first calculated the number of samples in each of the four classes, $\left \{n_1, n_2, \ldots, n_C\right \}$, and set a target sample size $N_{\text{target}} = \max(n_1, n_2, \ldots, n_C)$ to equalize class sizes. To achieve this, we duplicated samples from the minority classes until each class reached $N_{\text{target}}$. This process resulted in a balanced dataset, $\hat{D} = \left \{(x'_1, y'_1), (x'_2, y'_2), \ldots, (x'_M, y'_M)\right \}$, with $M = C \cdot N_{\text{target}}$ samples, ensuring an equal number of samples for each class.

After balancing the data, we created a dataset $\hat{D} = \left \{(x'_1, y'_1), (x'_2, y'_2), \ldots, (x'_M, y'_M)\right \}$ and employed a multilayer perceptron (MLP) \cite{bib35} to evaluate OSAHS severity. The MLP, a type of feedforward neural network, is known for its strong ability to fit nonlinear data. We developed an MLP model with LL hidden layers, where each layer's output is computed by applying an activation function to the weighted sum of its inputs.
\begin{equation}
z^{l}=W^{l} \cdot a^{l-1}+b^{l}
\end{equation}
\begin{equation}
a^{l}=\sigma(z^{l}) 
\end{equation}
Here, $W^{l}$ and $b^{l}$ are the weight matrix and bias vector for the $l$-th layer, respectively, while $a^{l-1}$ stands for the output from the preceding layer, and $\sigma$ denotes the activation function. For the output layer (the $L$-th layer), we utilized the softmax activation function to normalize the outputs, resulting in the final output $a^{L}$ for each sample.

Subsequently, we trained the model with an ordinal regression loss function, where each output $a^{L}$ has a predicted probability distribution $p_i = [p_{i1}, p_{i2}, \ldots, p_{iC}]$ for $C$ classes. The true label $y_i \in \{1, 2, \ldots, C\}$ is used, and the ordinal regression objective function is employed to consider the order relationship between the labels:
\begin{equation}
L_i = - \sum_{j=1}^{C} \left( \log P(y_i \leq j | a_{i}^{L}) + \log P(y_i > j | a_{i}^{L}) \right)
\end{equation}
$P(y_i \leq j | a_{i}^{L})$ represents the probability that the output $a^{L}$ predicts class $j$ or below, and $P(y_i > j | a_{i}^{L})$ represents the probability of predicting class $j$ or above. Minimizing the ordinal regression objective function $L_{i}$ helps the model learn the correct order relationships, enhancing its accuracy in detecting OSAHS severity.

\section{Experiments AND RESULTS}
This study analyzed a dataset of 500 OSAHS patients from a university hospital, including their facial images and OSAHS severity levels determined by three experts. The data were collected using standardized protocols to ensure accuracy. Key factors correlated with OSAHS diagnosis and easily obtainable during routine hospital visits, such as gender, age, neck circumference, BMI, WHR, and conditions like hypertension, diabetes, heart disease, and hyperlipidemia, were examined. 

The apnea-hypopnea index (AHI) \cite{bib63} is used to assess the severity of sleep disordered breathing by measuring the number of apnea and hypopnea events per hour of sleep. Higher AHI values indicate more severe conditions \cite{bib36}. Table \ref{tab1} shows the distribution of OSAHS patient severity levels based on AHI.
\begin{table}[htbp]
  \centering
  \caption{AHI-Based Severity Distribution of OSAHS Patients}\label{tab1}
  \begin{tabular}{*{5}{c}}  
    \toprule
    \textbf{Type} & \textbf{AHI} & \textbf{Number} & \textbf{Age(mean)} & \textbf{BMI(mean)} \\
    \midrule
Normal    & [0,5)   & 58  & 32.1  & 24\\
Mild    & [5,15)   & 76  & 39.1  & 25.1\\
Moderate    & [15,30)   & 76  & 43.2  & 26.6\\
Severe    & (30,+$\infty$)   & 290  & 41.4  & 28.6  \\
    \bottomrule
  \end{tabular}
\end{table}
\subsection{Experiment Setup} 
The experiment utilized an RTX4090 GPU and the PyTorch framework to build the model, optimized with the Adam optimizer \cite{bib37} and an adaptive learning rate based on validation performance. A dropout rate of 0.4 was used to prevent overfitting. The dataset was divided into training (80\%), validation (10\%), and test (10\%) sets, with the validation set for hyperparameter tuning and the test set for evaluating final model performance.

We evaluated the effectiveness of our approach using standard metrics for multi-class classification in deep learning, including accuracy, precision, recall, F1 score, AUC and variance. Weighted averaging was used for precision, recall, and F1 to account for class size. To handle data imbalance, we also incorporated the area under the precision-recall curve (AUPR) and accuracy metrics for each severity level. These metrics provided a comprehensive assessment of our method's performance.
 \begin{table*}[htbp]
  \centering
  \caption{The OSAHS severity diagnosis accuracy of our proposed method and the baseline methods.}\label{tab2}
  \begin{tabular}{ccccccccccc}
    \toprule
    \textbf{Model} & \textbf{Acc(\%)}  & \textbf{Pre(\%)}& \textbf{Rec(\%)}& \textbf{F1(\%)}&  \textbf{AUC(\%)}&  \textbf{Var(\%)}&  \textbf{Normal(\%)}&  \textbf{Mild(\%)}&  \textbf{Moderate(\%)}&  \textbf{Severe(\%)}\\
    \\
     &  & & & & &  &  \textbf{AUPR$|$Acc}&  \textbf{AUPR$|$Acc}&  \textbf{AUPR$|$Acc}&  \textbf{AUPR$|$Acc}\\
    \midrule
        ConvNeXt \cite{bib52} &  67.6 & 70.6 & 67.7& 66.3& 77.5&1.1&43.6$|$39.6&56.7$|$83.3&50.3$|$54.6&82.3$|$95.1\\
        CLIP \cite{bib53} & 73.7 & 76.5& 73.7 & 72.7& 84.3&0.9&49.6$|$41.4&63.2$|$77.1&65.8$|$82.8&91.8$|$91.9\\
        Florence-2 \cite{bib6}   & 79.7   & 80.6& 79.7& 79.4& 84.5 &0.9&54.4$|$\textbf{72.4}&61.1$|$72.9&70$|$71.8&91.1$|$\textbf{100} \\
        \textbf{VTA-OSAHS}  & \textbf{91.3}  & \textbf{92.1}& \textbf{91.3} & \textbf{91.1} & \textbf{95.6}&\textbf{0.7}&\textbf{80.9$|$72.4}&83.7$|$\textbf{97.9}&\textbf{90.6$|$95.3}&\textbf{98.9}$|$\textbf{100} \\
    \bottomrule
  \end{tabular}
\end{table*}

\subsection{OSAHS Severity Classification Results}
To assess the effectiveness of our proposed model for classifying OSAHS severity, we compared it with existing models, including ConvNeXt \cite{bib52}, CLIP \cite{bib53}, and Florence-2 \cite{bib6}. Florence-2, multimodal model, utilizes a prompt based approach and a sequence to sequence framework to convert input images into visual token embeddings that capture spatial and semantic information, which are then combined with text embeddings for further analysis.

Table \ref{tab2} shows that our proposed framework effectively diagnoses OSAHS severity, achieving a 91.3\% overall accuracy with a 0.7\% variance. The AUPR scores for severity levels range from 80.9\% for normal to 98.9\% for severe cases. In comparison, the ConvNeXt model achieves 67.6\% top-1 accuracy, while Florence-2 reaches 79.7\%. These results indicate that models relying only on image encoding, like ConvNeXt, fail to capture critical facial features for OSAHS, and Florence-2's performance is influenced by its reliance on a diverse pretraining dataset, which differs from our focused dataset of similar facial images.

The exceptional performance of our framework is due to its focus on annotating specific facial features relevant to OSAHS, such as neck thickness, a flattened facial profile, and a retruded mandible, rather than using traditional image classification methods. This targeted approach allows for a deeper analysis of facial images, extracting features that are crucial for accurate OSAHS severity prediction, and aligns better with clinical decision making requirements.
\begin{table}[htbp]
  \caption{The Impact of Different Modules}
  \label{tab_ablation}
  \centering
    \begin{tabular}{@{}p{2.5cm} l cccc c p{1cm} @{}}
    \toprule
    \textbf{Different Impact} & \textbf{Model} & \textbf{Acc(\%)} &  \textbf{AUC(\%)}\\
    \midrule
    \multirow{2}{*}{{Data Oversampling}} &
     {W/o.} Oversampling & 58.5  & 54.3 \\
     {}& SMOTE & {89.2} & {93.5} \\
    {}& {\bfseries ROS } & \textbf{91.3}  & \textbf{95.6}  \\
    \midrule
    \multirow{2}{*}{{Objective Function}} &
    CrossEntropy Loss & 88.6 & 94.6\\
    {} & {\bfseries Ordinal Regression } & \textbf{91.3}  & \textbf{95.6}  \\
    \midrule
    \multirow{2}{*}{{Modality Fusion}} &
    AutoEncoder & 87.9 & 92.3\\
    {} & {\bfseries Cross Attention } & \textbf{91.3}  & \textbf{95.6}  \\
  \midrule
  \multirow{2}{*}{{Different Modality}} &
    {Textual }  & 46.1  & 54.1\\
    {}&Visual & {85.7} & 89.3\\
    {}&\textbf{Multimodal } & \textbf{91.3} & \textbf{95.6}  \\
    \bottomrule
  \end{tabular}
\end{table} 
\subsection{Ablation Studies}
In this section, we performed a series of ablation experiments to evaluate the influence of each component within the proposed model framework. The findings from these ablation studies are presented in Table \ref{tab_ablation}.

\subsubsection{The effect of oversampling}
Applying random oversampling improved our model's accuracy to 91.3\% for classifying four OSAHS severity levels, slightly better than using SMOTE and significantly higher than the 58.5\% accuracy without oversampling. This improvement is due to addressing the substantial class imbalance in the real-world clinical dataset, which, without oversampling, can lead to poor training and classification of minority classes. The ROS algorithm effectively balanced the classes, enhancing diagnostic accuracy.

\subsubsection{The effect of different objective functions}
The experimental results show that using an ordinal regression objective function improves the model's performance in classifying OSAHS severity. Unlike the cross-entropy function, which is less suited for ordered data, the ordinal regression function considers the ordered nature of clinical data labels and maintains the sequence relationship between classes, thereby enhancing diagnostic accuracy.

\subsubsection{The effect of different data fusion methods}
The experimental results show that fusion strategies, specifically using cross-attention to focus on relevant image regions based on textual data, improve the model's performance in OSAHS severity classification. This approach avoids information loss or redundancy from directly merging modalities, allowing the model to capture richer contextual information and enhance its reasoning abilities. These findings highlight the importance of the cross-attention mechanism for effective feature integration in the VTA-OSAHS framework.

\subsubsection{The effect of different modalities}
The experimental results show that using multimodal data, combining images and text, enhances the diagnosis of OSAHS severity. Image data alone provided higher accuracy than text, as it captures crucial facial features needed for diagnosis. However, text-based methods struggle to interpret complex clinical relationships without specific context. By integrating both text and image data, multimodal approaches provide complementary information, improving the model's accuracy and reliability in detecting OSAHS severity.

\section{Conclusion}
This study presents \textbf{VTA-OSAHS}, an end-to-end, multi-modal framework for diagnosing OSAHS severity. The method uses an attention mesh for image feature extraction, Clinical BERT for text encoding, an attention mechanism for integrating modalities, the ROS for balancing samples, and an MLP with ordinal regression for classification. Tested on a hospital dataset, the framework shows potential for improving diagnostic accuracy and clinical applications. Future work will focus on model optimization and expanding the dataset to include more comprehensive clinical assessments and physiological data specific to OSAHS patients.

\end{document}